\def\a{\alpha}
\def\b{\beta}
\def\d{\delta}
\def\g{\gamma}

\def\l{\lambda}

\def\t#1{\mbox{\it #1}}
\def\ss#1{\mbox{\scriptsize #1}}

\def\ie{{\it i.e.}}
\def\eg{{\it e.g.}}

\def\bi{\begin{itemize}}
\def\ei{\end{itemize}}

\input{psfig}
\def\ss#1{_{\mbox{\scriptsize #1}}}

\documentstyle[aclap]{article}

\title{\vspace{-75pt}
{\normalsize \tt \hfill To appear in {\it Proceedings of the 34th Annual
Meeting of the ACL}, June 1996} \\ \mbox{} \\
An Empirical Study of Smoothing Techniques for Language Modeling}

\author{
Stanley F. Chen \\
Harvard University\\
Aiken Computation Laboratory\\
33 Oxford St. \\
Cambridge, MA 02138\\
{\tt sfc@eecs.harvard.edu}
\And
Joshua Goodman\\
Harvard University\\
Aiken Computation Laboratory\\
33 Oxford St. \\
Cambridge, MA 02138\\
{\tt goodman@eecs.harvard.edu}
}

\begin{document}

\maketitle
\vspace{-0.5in}
\begin{abstract}
We present an extensive empirical comparison of several smoothing
techniques in the domain of language modeling,
including those described by Jelinek and Mercer~(1980),
Katz~(1987), and Church and Gale~(1991).
We investigate for the first time
how factors such as training data size, corpus (\eg, Brown
versus Wall Street Journal), and $n$-gram order (bigram versus trigram)
affect the relative performance of these methods, which we measure
through the cross-entropy of test data.  In addition, we introduce
two novel smoothing techniques, one a variation of Jelinek-Mercer
smoothing and one a very simple linear interpolation technique, both
of which outperform existing methods.
\end{abstract}

\section{Introduction}

{\it Smoothing\/} is a technique essential in the construction of $n$-gram
language models, a staple in speech recognition \cite{Bahl:83a}
as well as many other domains
\cite{Church:88a,Brown:90b,Kernighan:90a}.
A {\it language model\/}
is a probability distribution over strings $P(s)$ that attempts
to reflect the frequency with which each string $s$ occurs as a sentence
in natural text.  Language models are used in speech recognition
to resolve acoustically ambiguous utterances.  For example, if we have that
$P(\mbox{\it it takes two}) \gg P(\mbox{\it it takes too})$, then we know
{\it ceteris paribus\/} to prefer the former transcription over the latter.

While smoothing is a central issue in language modeling, the
literature lacks a definitive comparison between the many
existing techniques.
Previous studies \cite{Nadas:84a,Katz:87a,Church:91a,MacKay:95a}
only compare a small number of methods (typically two) on
a single corpus and using a single training data size.  As a result,
it is currently difficult for a researcher to intelligently choose
between smoothing schemes.

In this work, we carry out an extensive empirical comparison of
the most widely used smoothing techniques, including those
described by \newcite{Jelinek:80a}, \newcite{Katz:87a},
and Church and Gale (1991).  We carry out
experiments over many training data sizes on varied
corpora using both bigram and trigram models.  We demonstrate that
the relative performance of techniques depends greatly on
training data size and $n$-gram order.  For example, for bigram models
produced from large training sets
Church-Gale smoothing has superior performance, while Katz smoothing
performs best on bigram models produced from smaller data.
For the methods with parameters that can be tuned to improve performance,
we perform an automated search for optimal values and show that
sub-optimal parameter selection can significantly decrease performance.
To our knowledge, this is the first smoothing work that systematically
investigates any of these issues.

In addition, we introduce two novel smoothing techniques:
the first belonging to
the class of smoothing models described by Jelinek and Mercer,
the second a very simple linear interpolation method.  While being
relatively simple to implement, we show
that these methods yield good performance in bigram models and
superior performance in trigram models.

We take the performance of a method $m$ to be its
cross-entropy on test data
$$ \frac{1}{N_T} \sum_{i=1}^{l_T} -\log_2 P_m(t_i) $$
where $P_m(t_i)$ denotes the language model produced with method $m$ and
where the test data $T$ is composed of sentences $(t_1, \ldots, t_{l_T})$
and contains a total of $N_T$ words.  The entropy is inversely
related to the average probability a model assigns to sentences in
the test data, and it is generally assumed
that lower entropy correlates with better performance in applications.

\subsection{Smoothing $n$-gram Models}

In $n$-gram language modeling, the probability of a string $P(s)$ is
expressed as the product of the probabilities of the words that
compose the string, with each word probability conditional on
the identity of the last $n - 1$ words,
\ie, if $s = w_1\cdots w_l$ we have
\begin{equation}
P(s) = \prod_{i=1}^l P(w_i | w_1^{i-1}) \approx
	\prod_{i=1}^l P(w_i | w_{i-n+1}^{i-1}) \label{eqn:sent}
\end{equation}
where $w_i^j$ denotes the words $w_i\cdots w_j$.
Typically, $n$ is taken to be two or three, corresponding
to a {\it bigram\/} or {\it trigram\/} model, respectively.\footnote{To make
the term $P(w_i | w_{i-n+1}^{i-1})$ meaningful for $i < n$, one can
pad the beginning of the string with a distinguished token.  In this
work, we assume there are $n-1$ such distinguished tokens preceding each
sentence.}

Consider the case $n=2$.  To estimate the probabilities
$P(w_i | w_{i-1})$ in equation (\ref{eqn:sent}), one can acquire
a large corpus of text, which we refer to as {\it training data},
and take
\begin{eqnarray*}
P\ss{ML}(w_i | w_{i-1})  &= & \frac{P(w_{i-1} w_i)}{P(w_{i-1})}  \\
& = & \frac{c(w_{i-1} w_i) / N_S}{c(w_{i-1}) / N_S} \\
& = & \frac{c(w_{i-1} w_i)}{c(w_{i-1})} 
\end{eqnarray*}
where $c(\a)$ denotes the number of times the string $\a$ occurs in the text
and $N_S$ denotes the total number of words.
This is called the {\it maximum likelihood\/} (ML) estimate for
$P(w_i | w_{i-1})$.


While intuitive, the maximum likelihood estimate is a poor one when the
amount of training data is small compared to the size of the model
being built, as is generally the case in language modeling.  For example,
consider the situation where a pair of words, or {\it bigram\/},
say {\it burnish the}, doesn't occur in the training data.  Then,
we have $P\ss{ML}(\t{the} | \t{burnish}) = 0$, which is clearly inaccurate
as this probability should be larger than zero.  A zero
bigram probability
can lead to errors in speech recognition, as it disallows the bigram
regardless of how informative the acoustic signal is.
The term {\it smoothing\/}
describes techniques for adjusting the maximum likelihood
estimate to hopefully produce more accurate probabilities.

As an example, one simple smoothing technique is to pretend each bigram
occurs once more than it actually did
\cite{Lidstone:20a,Johnson:32a,Jeffreys:48a}, yielding
$$ P_{+1}(w_i | w_{i-1}) = \frac{c(w_{i-1} w_i) + 1}{c(w_{i-1}) + |V|} $$
where $V$ is the vocabulary, the set of all words being considered.
This has the desirable
quality of preventing zero bigram probabilities.  However, this scheme
has the flaw of assigning the same probability to say,
{\it burnish the\/} and {\it burnish thou\/} (assuming neither occurred
in the training data), even though intuitively the former seems
more likely because the word {\it the\/} is much more common than
{\it thou}.

To address this, another smoothing technique is to
{\it interpolate\/} the bigram model with
a unigram model $P\ss{ML}(w_i) = c(w_i) / N_S$, a model that
reflects how often each word
occurs in the training data.  For example, we can take
$$ P\ss{interp}(w_i | w_{i-1}) =
	\l P\ss{ML}(w_i | w_{i-1}) + (1 - \l) P\ss{ML}(w_i) $$
getting the behavior that bigrams involving common words are
assigned higher probabilities \cite{Jelinek:80a}.

\begin{figure*}[t]
$$ \psfig{figure=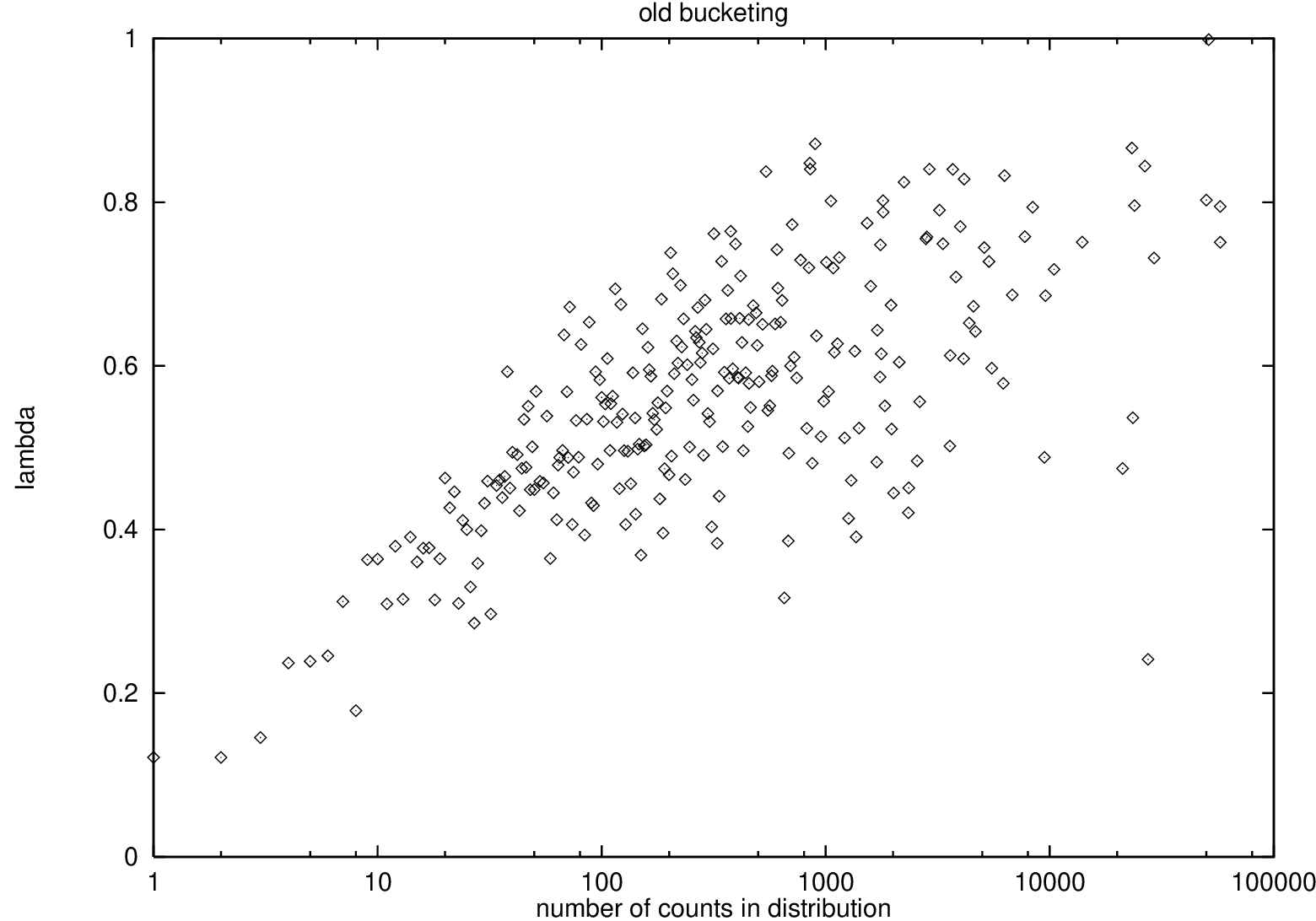,width=3in} \hspace{0.3in}
\psfig{figure=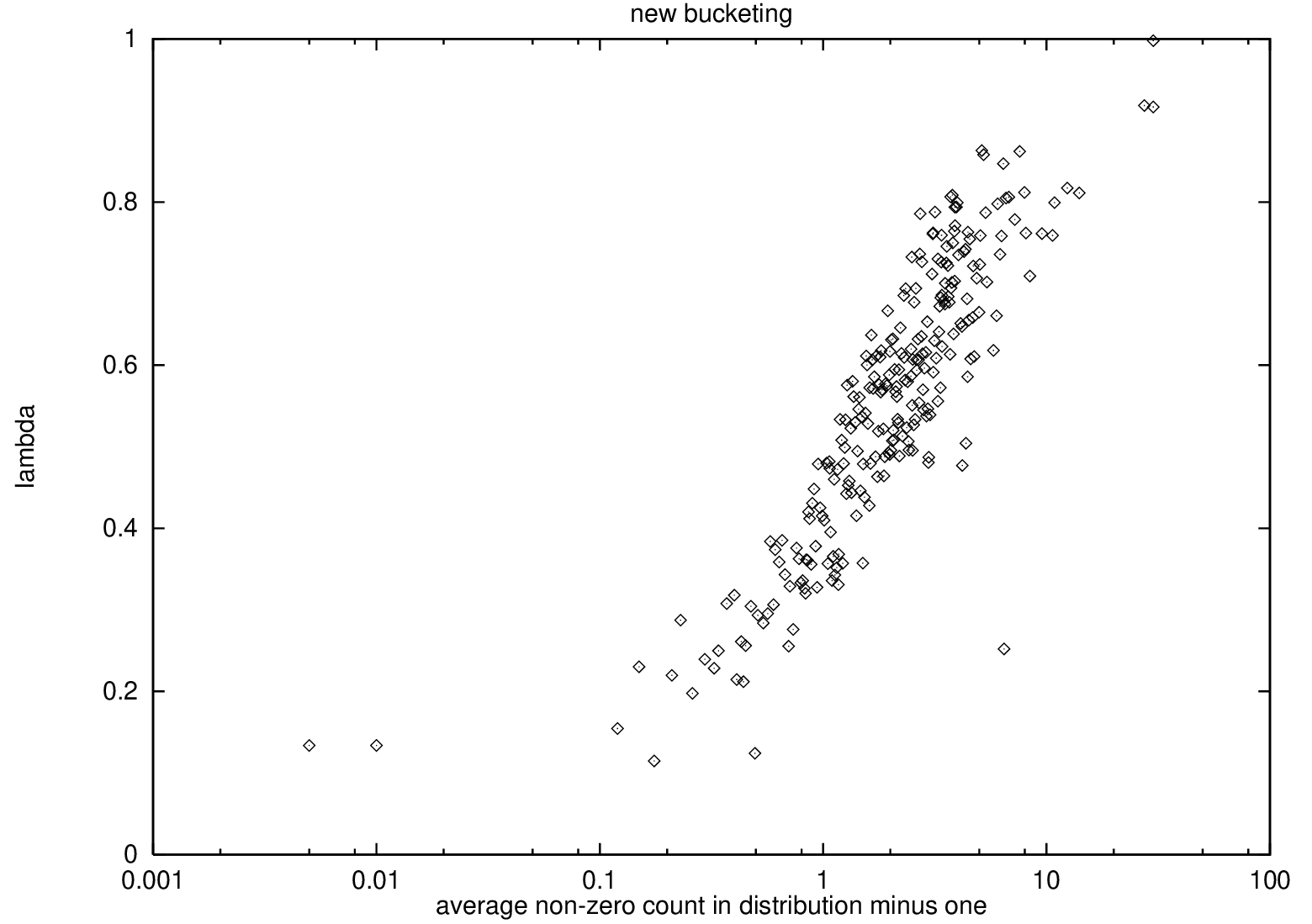,width=3in} $$
\caption{$\l$ values for old and new bucketing schemes for
Jelinek-Mercer smoothing; each point represents a single bucket}
	\label{fig:lamb}
\end{figure*}

\section{Previous Work}

The simplest type of smoothing used in practice is {\it additive\/}
smoothing \cite{Lidstone:20a,Johnson:32a,Jeffreys:48a}, where we take
\begin{equation}
P\ss{add}(w_i | w_{i-n+1}^{i-1}) =
	\frac{c(w_{i-n+1}^i) + \d}{c(w_{i-n+1}^{i-1}) + \d |V|} \label{eqn:add}
\end{equation}
and where Lidstone and Jeffreys advocate $\d = 1$. 
Gale and Church \shortcite{Gale:90a,Gale:94a}
have argued that this method generally performs poorly.

The Good-Turing estimate \cite{Good:53a} is central
to many smoothing techniques.  It is not used directly for
$n$-gram smoothing because, like additive smoothing, it does
not perform the interpolation of lower- and higher-order models essential
for good performance.
Good-Turing states that an $n$-gram that occurs $r$ times should
be treated as if it had occurred $r^*$ times, where
$$ r^* = (r + 1) \frac{n_{r+1}}{n_r} $$
and where $n_r$ is the number of
$n$-grams that occur exactly $r$ times in the training data.

Katz smoothing \shortcite{Katz:87a} extends the intuitions of Good-Turing
by adding the interpolation of higher-order models
with lower-order models.  It is perhaps the most widely used
smoothing technique in speech recognition.

\newcite{Church:91a} describe a smoothing method that
combines the Good-Turing estimate with {\it bucketing}, the technique
of partitioning a set of $n$-grams into disjoint groups, where each
group is characterized independently through a set of parameters.  Like Katz,
models are defined recursively in terms of lower-order models.  Each
$n$-gram is assigned to one of several buckets based on its frequency
predicted from lower-order models.  Each bucket is treated as a
separate distribution and Good-Turing estimation is performed within
each, giving corrected counts that are normalized to yield
probabilities.

The other smoothing technique besides Katz smoothing
widely used in speech recognition
is due to \newcite{Jelinek:80a}.  They present
a class of smoothing models that involve linear interpolation, \eg,
\newcite{Brown:91h} take
\begin{eqnarray}
\lefteqn {P\ss{interp}(w_i|w_{i-n+1}^{i-1}) =}  & & \nonumber \\
 & &	\l_{w_{i-n+1}^{i-1}}\; P\ss{ML}(w_i | w_{i-n+1}^{i-1}) + \nonumber \\
 & &	(1 - \l_{w_{i-n+1}^{i-1}})\; P\ss{interp}(w_i |w_{i-n+2}^{i-1})
	\label{eqn:interp}
\end{eqnarray}
That is, the maximum likelihood estimate is interpolated with the smoothed
lower-order distribution, which is defined analogously.  Training
a distinct $\l_{w_{i-n+1}^{i-1}}$ for each $w_{i-n+1}^{i-1}$ is not
generally felicitous;
\newcite{Bahl:83a} suggest partitioning the $\l_{w_{i-n+1}^{i-1}}$
into buckets according to $c(w_{i-n+1}^{i-1})$, where all $\l_{w_{i-n+1}^{i-1}}$
in the same bucket are constrained to have the same value.

To yield meaningful results, the data used to estimate the
$\l_{w_{i-n+1}^{i-1}}$ need to be disjoint from the data used to
calculate $P\ss{ML}$.\footnote{When the same data is used to estimate
both, setting all $\l_{w_{i-n+1}^{i-1}}$ to one yields the optimal
result.}  In {\it held-out interpolation}, one reserves a section of
the training data for this
purpose.  Alternatively, Jelinek and Mercer describe a technique
called {\it deleted interpolation\/} where different parts of the
training data rotate in training either $P\ss{ML}$ or the
$\l_{w_{i-n+1}^{i-1}}$; the results are then averaged.

Several smoothing techniques are motivated within a Bayesian framework,
including work by \newcite{Nadas:84a} and \newcite{MacKay:95a}.

\section{Novel Smoothing Techniques}

Of the great many novel methods that we have tried, two techniques
have performed especially well.

\subsection{Method {\it average-count}} \label{sec:methoda}

This scheme is an instance of Jelinek-Mercer smoothing.  Referring
to equation (\ref{eqn:interp}), recall that
Bahl et al.\ suggest bucketing the $\l_{w_{i-n+1}^{i-1}}$
according to $c(w_{i-n+1}^{i-1})$.  We have found that partitioning
the $\l_{w_{i-n+1}^{i-1}}$ according to the average number of counts
per non-zero element
$\frac{c(w_{i-n+1}^{i-1})}{|w_i : c(w_{i-n+1}^i) > 0|}$
yields better results.

Intuitively, the less sparse the data for estimating
$P\ss{ML}(w_i | w_{i-n+1}^{i-1})$, the larger $\l_{w_{i-n+1}^{i-1}}$
should be.  While larger $c(w_{i-n+1}^{i-1})$ generally correspond
to less sparse distributions, this quantity ignores the allocation
of counts between words.  For example, we would consider a distribution
with ten counts distributed evenly among ten words to be much more sparse
than a distribution with ten counts all on a single word.  The
average number of counts per word seems to more directly express
the concept of sparseness.

In Figure \ref{fig:lamb}, we graph the value of $\l$
assigned to each bucket
under the original and new bucketing schemes on identical data.
Notice that the new bucketing scheme results in a much tighter plot,
indicating that it is better at grouping together distributions
with similar behavior.

\subsection{Method {\it one-count}}

This technique combines two intuitions.  First,
\newcite{MacKay:95a} argue that a reasonable form for a
smoothed distribution is
$$ P\ss{one}(w_i | w_{i-n + 1}^{i-1}) = \frac{c(w_{i-n +1}^{i}) +
	\a P\ss{one}(w_i | w_{i-n + 2}^{i-1})}{c(w_{i-n +1}^{i-1}) + \a} $$
The parameter $\a$ can be thought of as the number of counts being added
to the given distribution, where the new counts are distributed
as in the lower-order distribution.  Secondly, the Good-Turing
estimate can be interpreted as stating that the number of these extra
counts should be proportional to the number of words with
exactly one count in the given distribution.  We have found that taking
\begin{equation}
\a = \g\; [ n_1(w_{i-n+1}^{i-1}) + \b] \label{eqn:newb}
\end{equation}
works well, where $$n_1(w_{i-n+1}^{i-1}) = | w_i : c(w_{i-n+1}^i) = 1
|$$ is the number of words with one count, and where $\b$ and $\g$ are
constants.

\section{Experimental Methodology}

\subsection{Data}

We used the Penn treebank and TIPSTER corpora distributed
by the Linguistic Data Consortium.  From the treebank, we extracted
text from the tagged Brown corpus,
yielding about one million words.  From
TIPSTER, we used the Associated Press (AP), Wall Street Journal (WSJ), and
San Jose Mercury News (SJM) data, yielding 123, 84, and 43 million
words respectively.  We created two distinct vocabularies, one for the Brown
corpus and one for the TIPSTER data.  The former vocabulary
contains all 53,850 words occurring in Brown;
the latter vocabulary consists of the 65,173 words
occurring at least 70 times in TIPSTER.

For each experiment, we selected three segments of held-out data
along with the segment of training data.  One held-out segment was used as the
test data for performance evaluation, and the other two were
used as development test data
for optimizing the parameters of each smoothing method.
Each piece of held-out data was chosen to be
roughly 50,000 words.  This decision does not
reflect practice very well, as when the training data size is less
than 50,000 words it is not realistic to have so much
development test data available.  However, we made this decision to prevent
us having to optimize the training versus held-out data tradeoff for each
data size.  In addition, the development test
data is used to optimize typically
very few parameters, so in practice small held-out
sets are generally adequate, and perhaps can be avoided altogether
with techniques such as deleted estimation.

\subsection{Smoothing Implementations}

In this section, we discuss the details of our implementations of
various smoothing techniques.  Due to space limitations, these
descriptions are not comprehensive; a more
complete discussion is presented in \newcite{Chen:96b}.  The titles
of the following sections include the mnemonic we use to refer
to the implementations in later sections.  Unless otherwise
specified, for those smoothing models defined recursively in terms
of lower-order models, we end the recursion by taking the $n=0$
distribution to be the uniform distribution
$P\ss{unif}(w_i) = 1 / |V|$.  For each method, we
highlight the parameters (\eg, $\l_n$ and $\d$ below)
that can be tuned to optimize performance.  Parameter values
are determined through training on held-out data.

\subsubsection{Baseline Smoothing ({\tt interp-baseline})}

For our baseline smoothing method, we use an instance of
Jelinek-Mercer smoothing where we constrain all $\l_{w_{i-n+1}^{i-1}}$
to be equal to a single value $\l_n$ for each $n$, \ie,
\begin{eqnarray*}
P\ss{base}(w_i | w_{i-n+1}^{i-1}) & = &
	\l_n\; P\ss{ML}(w_i | w_{i-n+1}^{i-1}) + \\
& &	(1 - \l_n)\; P\ss{base}(w_i | w_{i-n+2}^{i-1})
\end{eqnarray*}

\subsubsection{Additive Smoothing ({\tt plus-one} and {\tt plus-delta})}

We consider two versions of additive smoothing.  Referring to
equation (\ref{eqn:add}), we fix $\d = 1$
in {\tt plus-one} smoothing.  In {\tt plus-delta}, we consider
any $\d$.

\subsubsection{Katz Smoothing ({\tt katz})}

While the original paper \cite{Katz:87a} uses a single parameter $k$,
we instead use a different $k$ for each $n > 1$, $k_n$.  We smooth the unigram
distribution using additive smoothing with parameter $\d$.

\subsubsection{Church-Gale Smoothing ({\tt church-gale})}

To smooth the counts $n_r$ needed for the Good-Turing
estimate, we use the technique described by \newcite{Gale:95a}.
We smooth the unigram distribution using Good-Turing without any
bucketing.

Instead of the bucketing scheme described in the
original paper, we use a scheme analogous to the one described
by \newcite{Bahl:83a}.  We make the assumption
that whether a bucket is large enough for accurate Good-Turing estimation
depends on how many $n$-grams with non-zero counts occur in it.
Thus, instead of partitioning the space of $P(w_{i-1}) P(w_i)$ values
in some uniform way as was done by Church and Gale, we partition
the space so that at least $c\ss{min}$ non-zero $n$-grams fall in each bucket.

Finally, the original paper describes only bigram smoothing in detail;
extending this method to trigram smoothing is ambiguous.  In particular,
it is unclear whether to bucket trigrams according to
$P(w_{i-2}^{i-1}) P(w_i)$ or $P(w_{i-2}^{i-1}) P(w_i | w_{i-1})$.
We chose the former; while the latter may yield better
performance, our belief is that it is much more difficult to implement and
that it requires a great deal more computation.

\subsubsection{Jelinek-Mercer Smoothing ({\tt interp-held-out} and
{\tt interp-del-int})}

We implemented two versions of Jelinek-Mercer smoothing differing
only in what data is used to train the $\l$'s.  We
bucket the $\l_{w_{i-n+1}^{i-1}}$ according to $c(w_{i-n+1}^{i-1})$
as suggested by Bahl et al.
Similar to our Church-Gale implementation, we choose
buckets to ensure that at least $c\ss{min}$ words in
the data used to train the $\l$'s fall in each bucket.

In {\tt interp-held-out}, the $\l$'s are trained using held-out interpolation
on one of the development test sets.
In {\tt interp-del-int}, the $\l$'s are trained
using the {\it relaxed deleted interpolation\/} technique described by
Jelinek and Mercer, where one word is deleted at a time.
In {\tt interp-del-int}, we bucket
an $n$-gram according to its count before deletion, as this turned out
to significantly improve performance.

\subsubsection{Novel Smoothing Methods ({\tt new-avg-count} and
	{\tt new-one-count})}

The implementation {\tt new-avg-count}, corresponding
to smoothing method {\it average-count}, is identical to
{\tt interp-held-out} except that we use the novel bucketing scheme described
in section \ref{sec:methoda}.
In the implementation {\tt new-one-count}, we
have different parameters $\b_n$ and $\g_n$ in equation (\ref{eqn:newb})
for each $n$.

%

\begin{figure*}[p]
$$ \psfig{figure=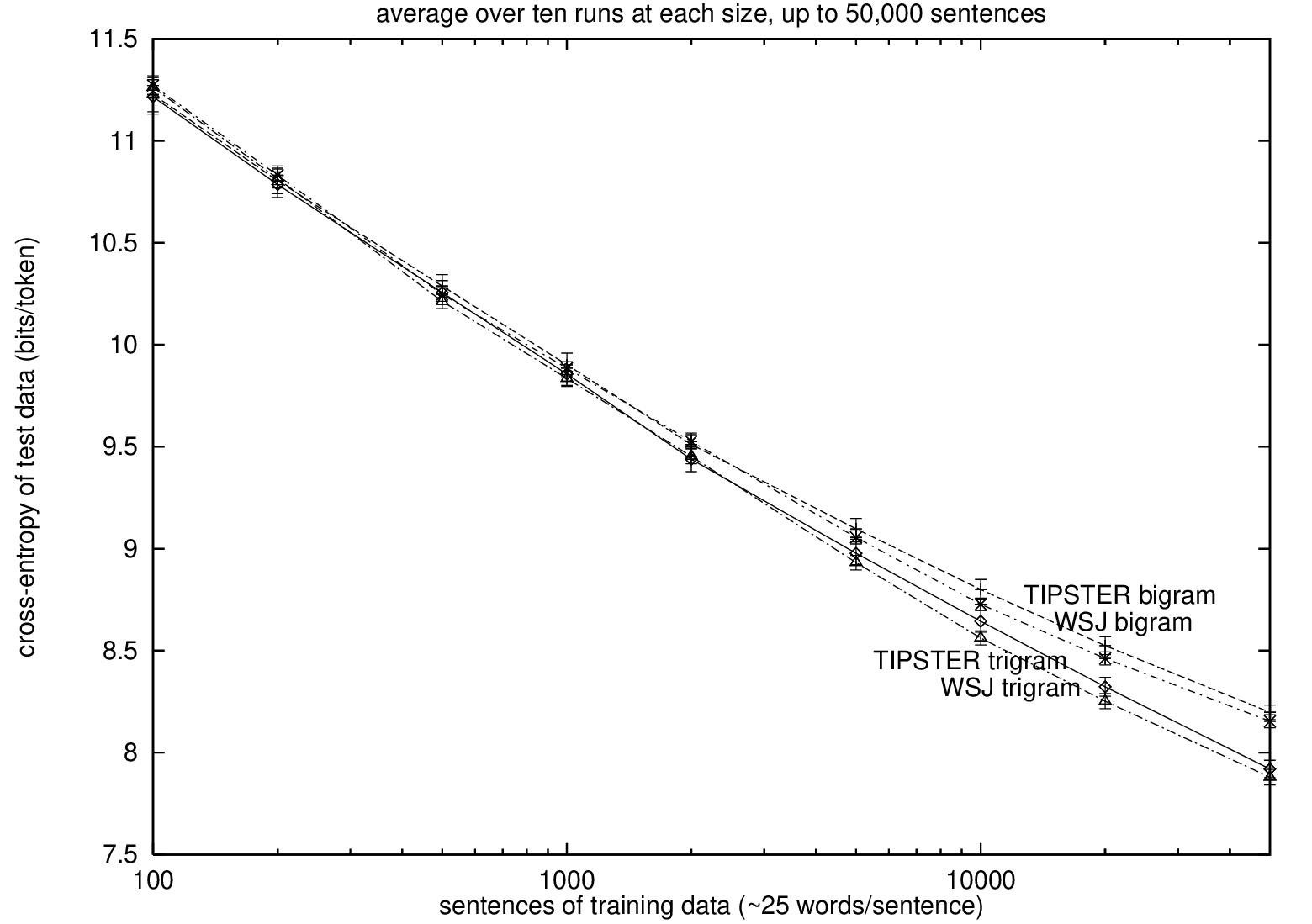,width=3in} \hspace{0.3in}
\psfig{figure=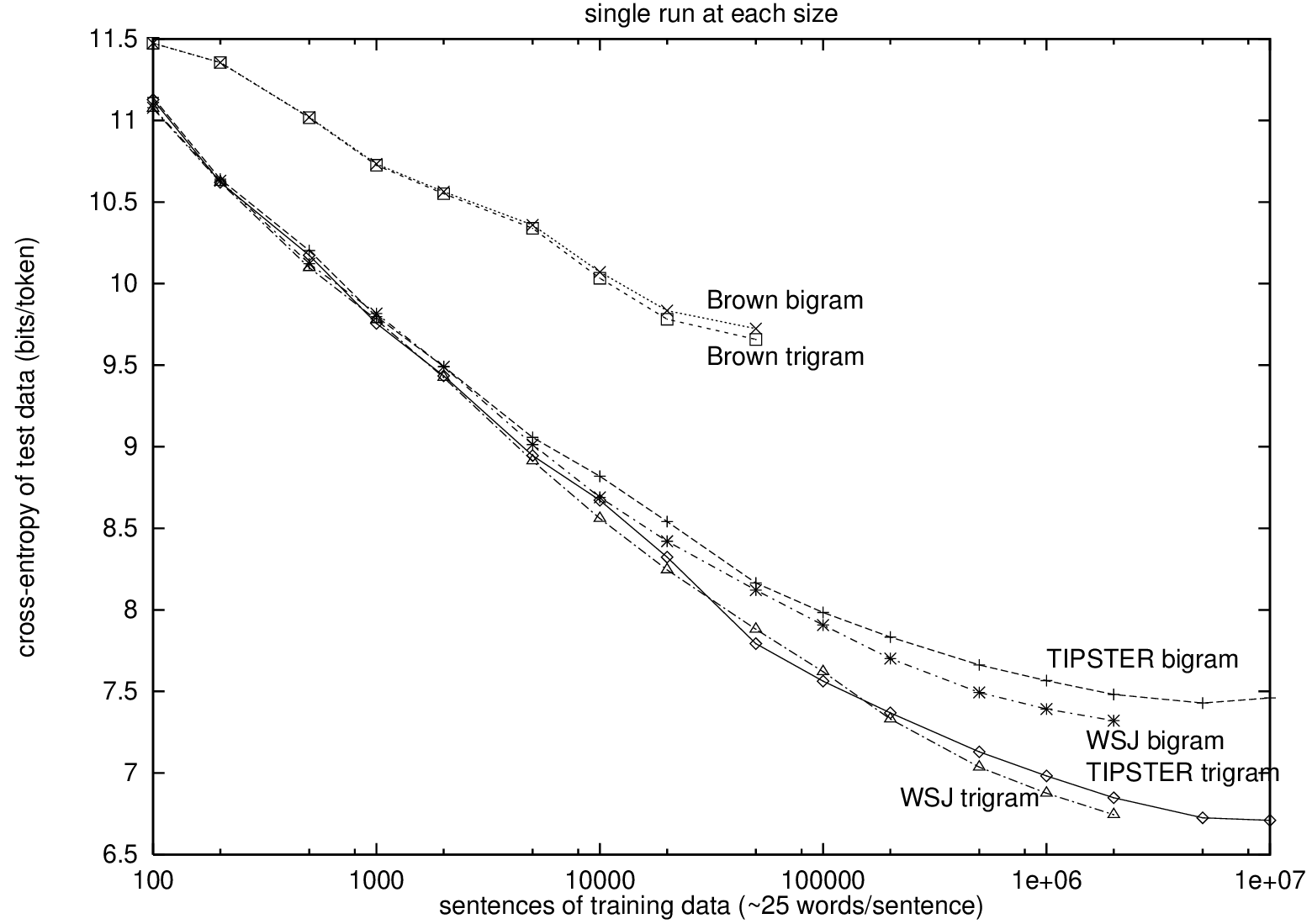,width=3in} $$
\caption{Baseline cross-entropy on test data; graph on left displays
averages over ten runs for training sets up to 50,000 sentences, graph
on right displays single runs for training sets up to 10,000,000 sentences}
	\label{fig:base}
\end{figure*}

\begin{figure*}[p]
$$ \psfig{figure=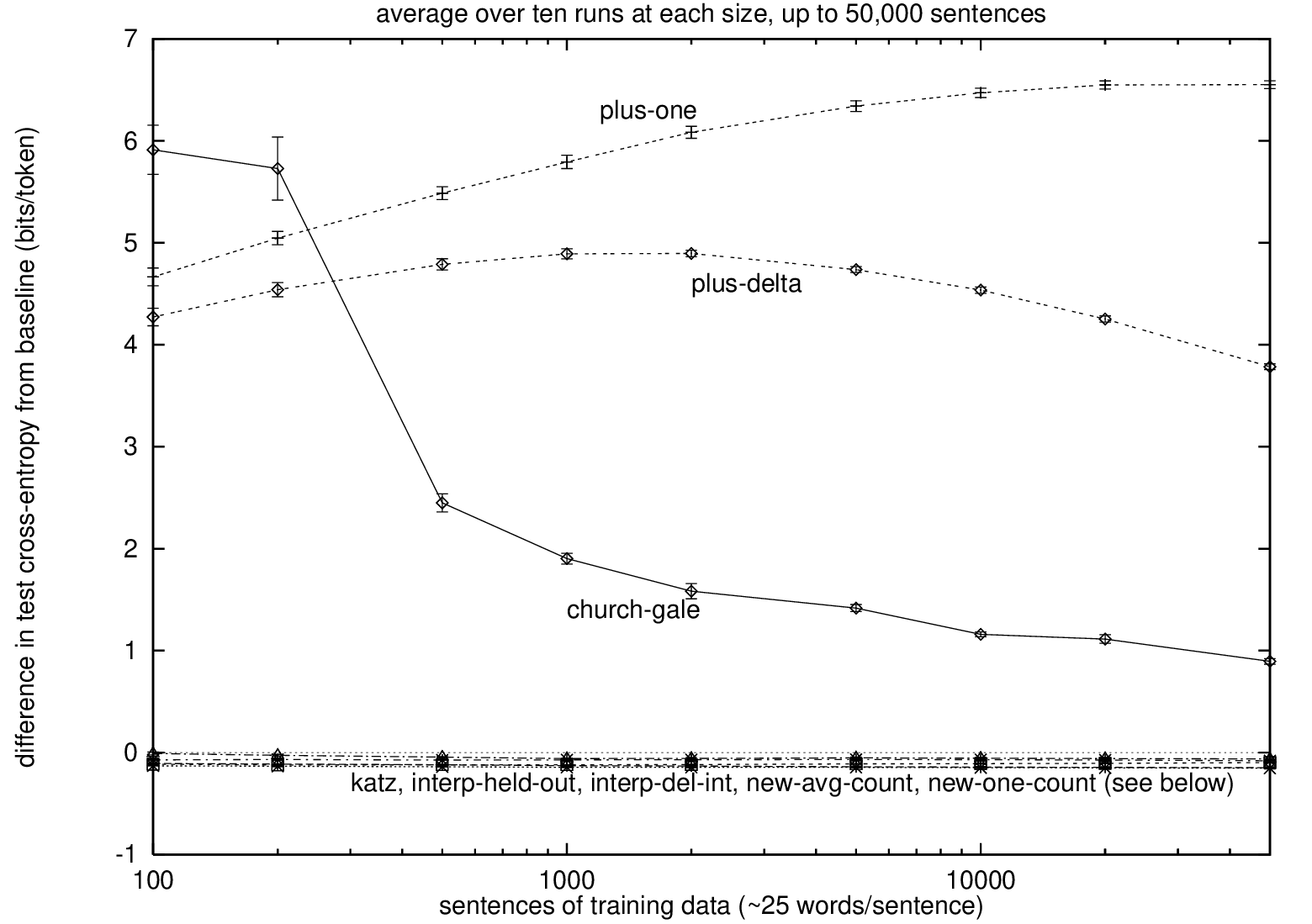,width=3in} \hspace{0.3in}
\psfig{figure=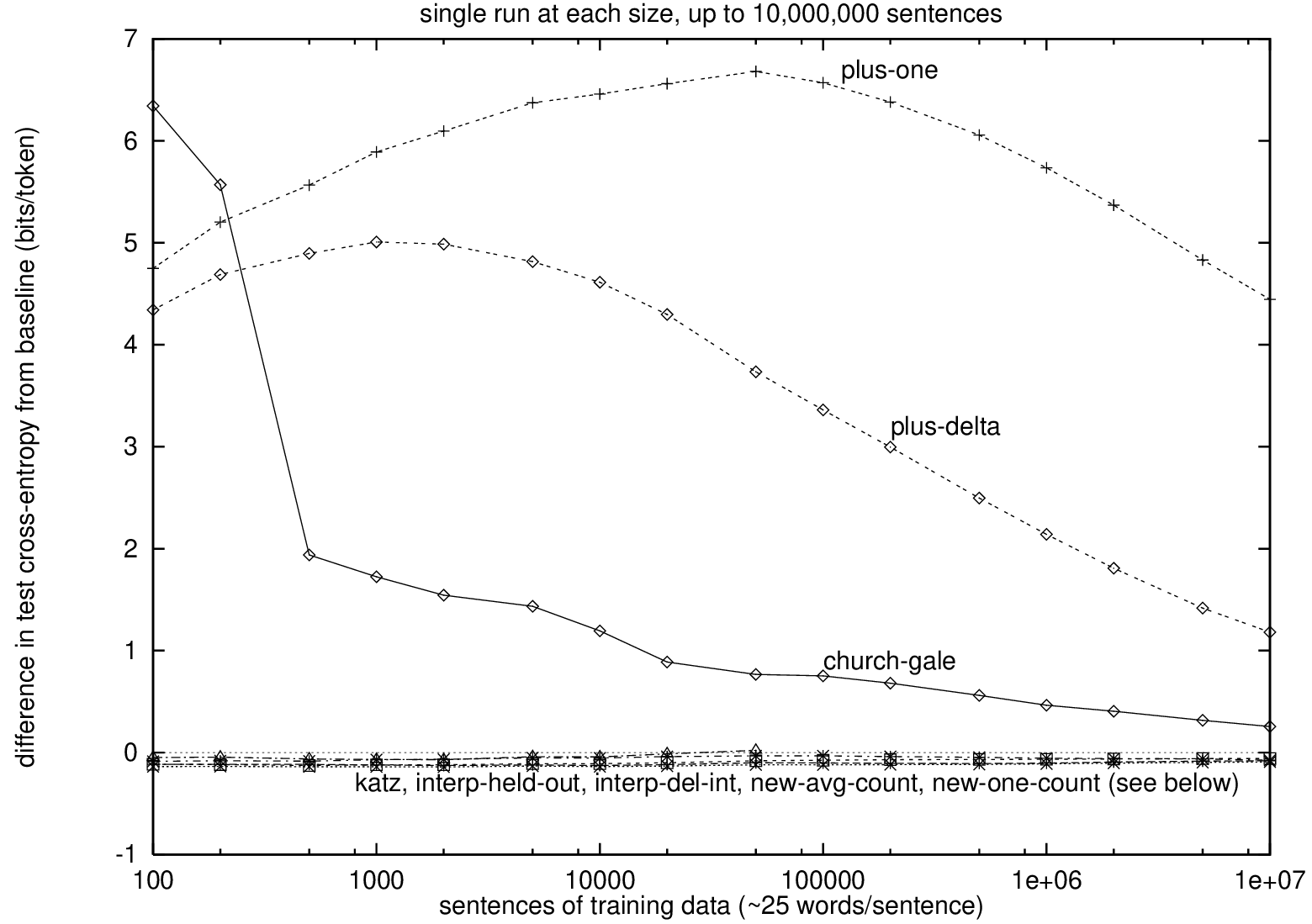,width=3in} $$
$$ \psfig{figure=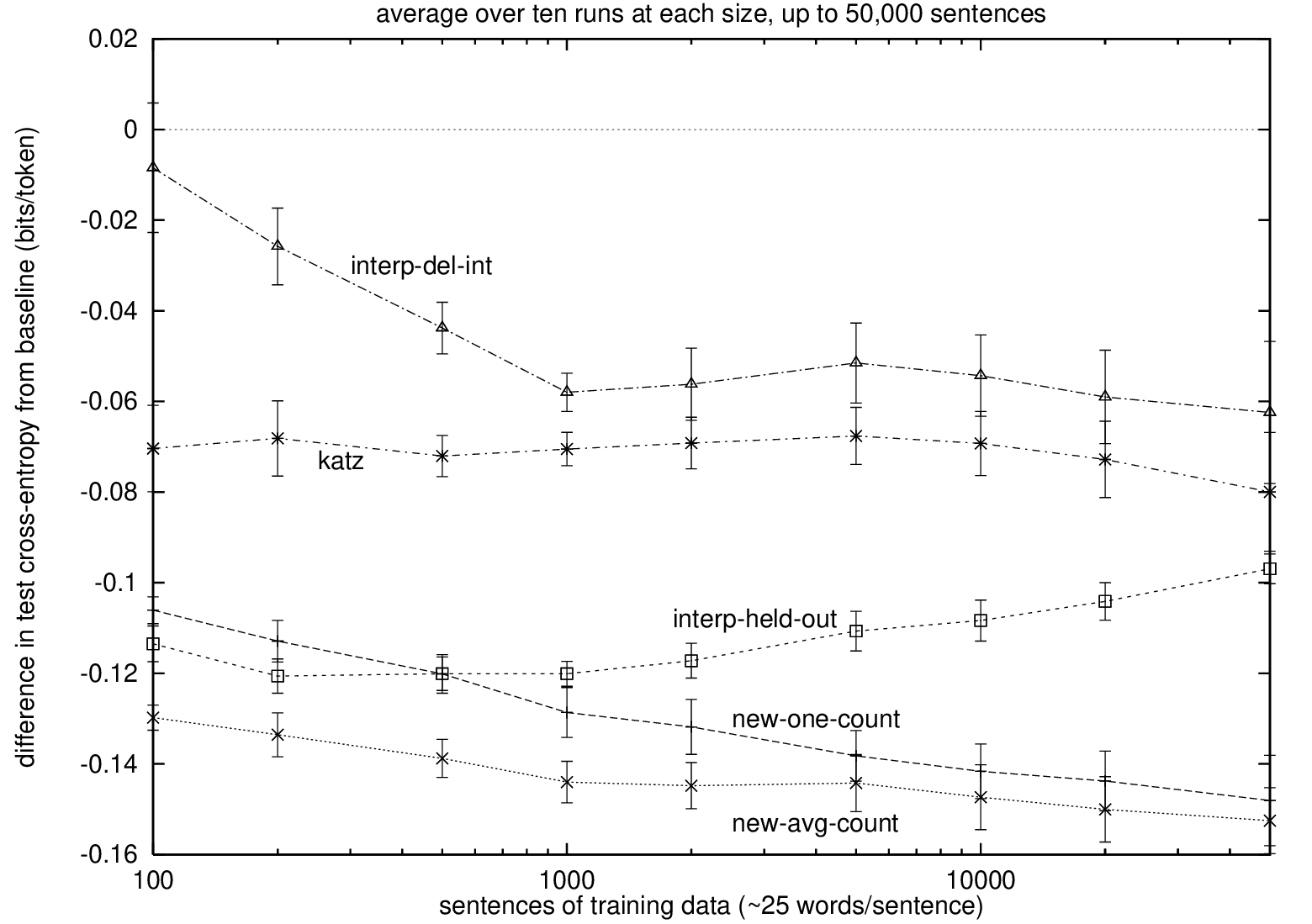,width=3in} \hspace{0.3in}
\psfig{figure=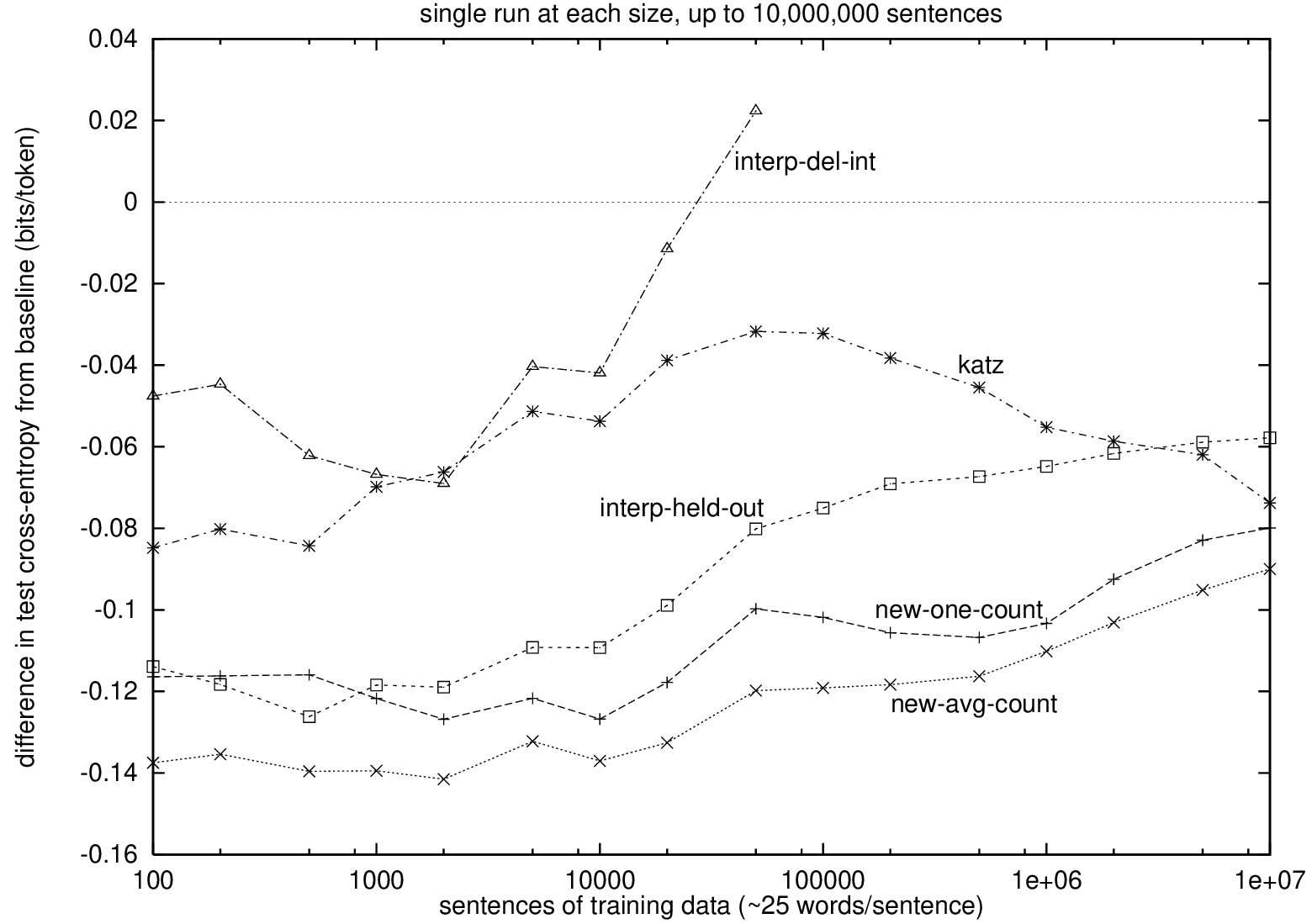,width=3in} $$
\caption{Trigram model on TIPSTER data;
	relative performance of various methods with respect
	to baseline; graphs on left display
	averages over ten runs for training sets up to 50,000 sentences, graphs
	on right display single runs for training sets up to 10,000,000 sentences;
	top graphs show all algorithms, bottom graphs zoom in on those methods
	that perform better than the baseline method}
	\label{fig:tip3}
\end{figure*}

\begin{figure*}[p]
$$ \psfig{figure=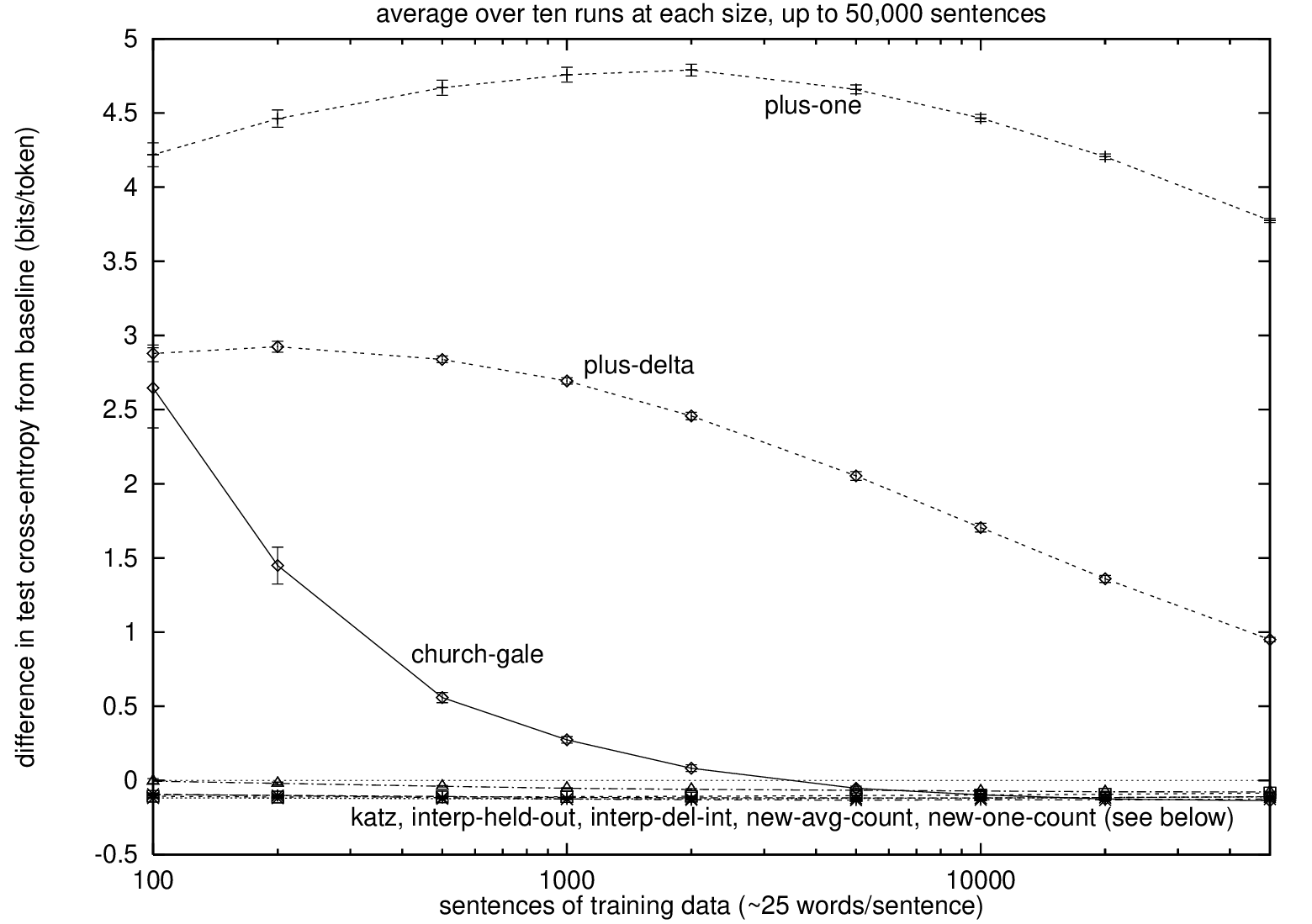,width=3in} \hspace{0.3in}
\psfig{figure=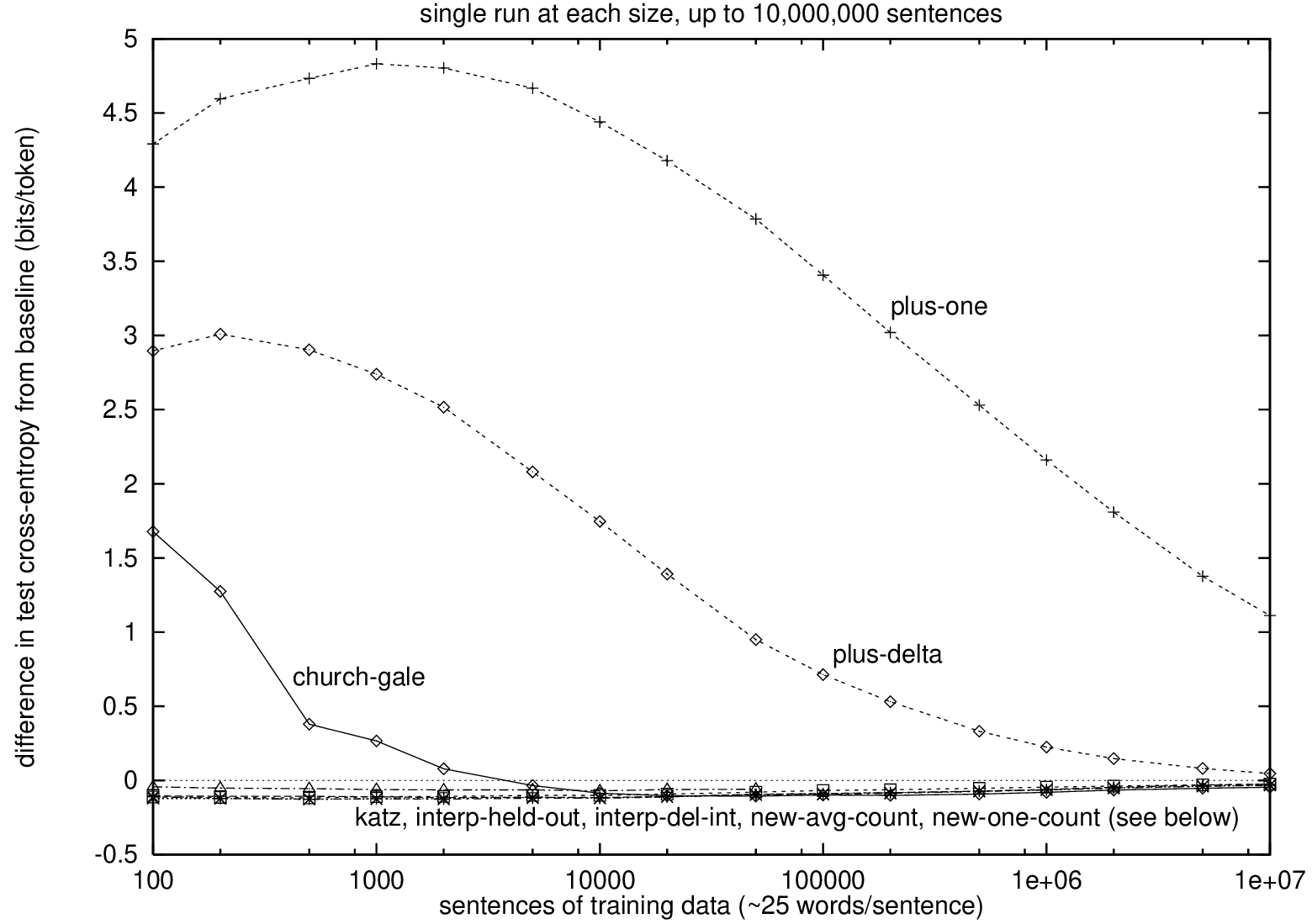,width=3in} $$
$$ \psfig{figure=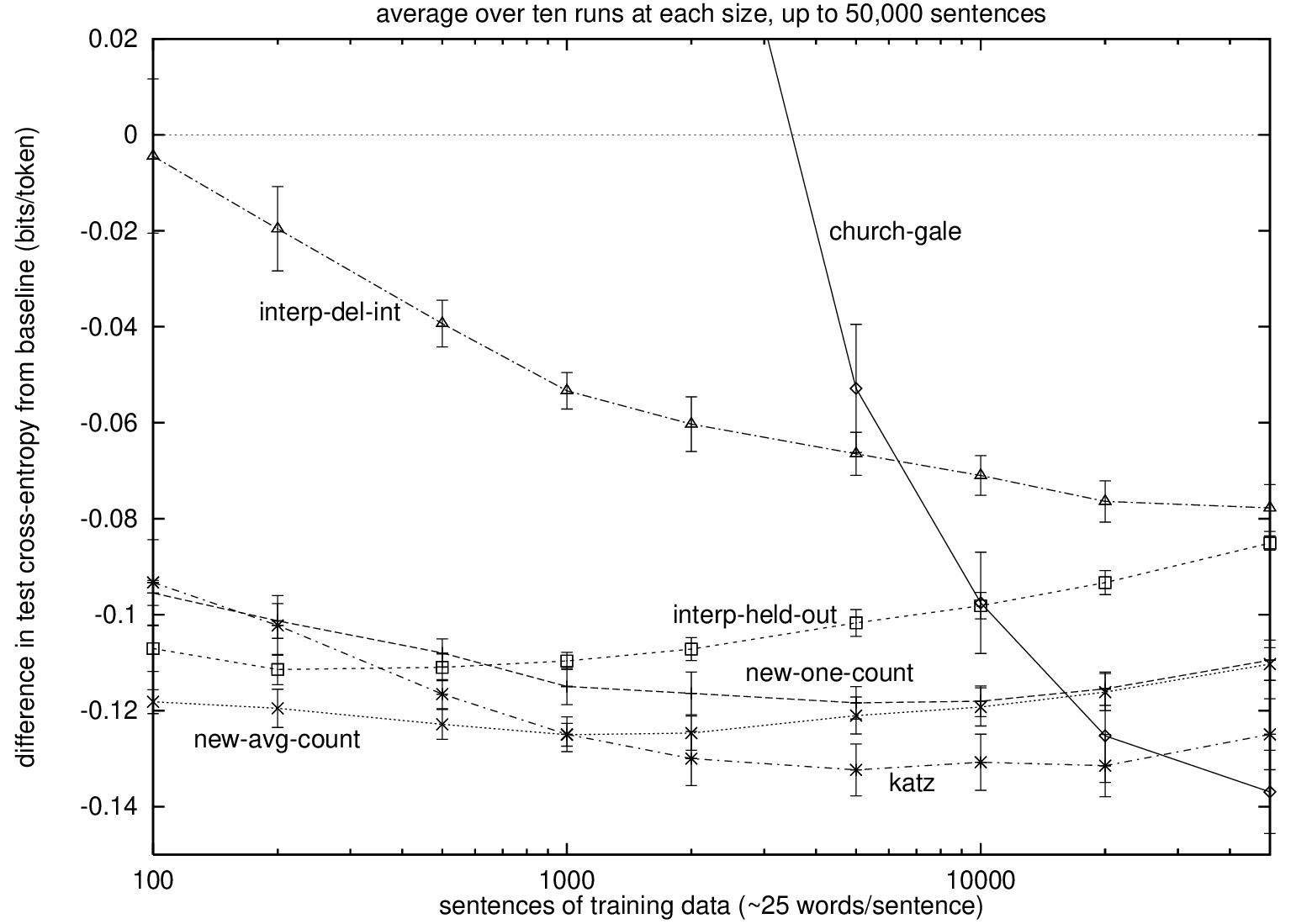,width=3in} \hspace{0.3in}
\psfig{figure=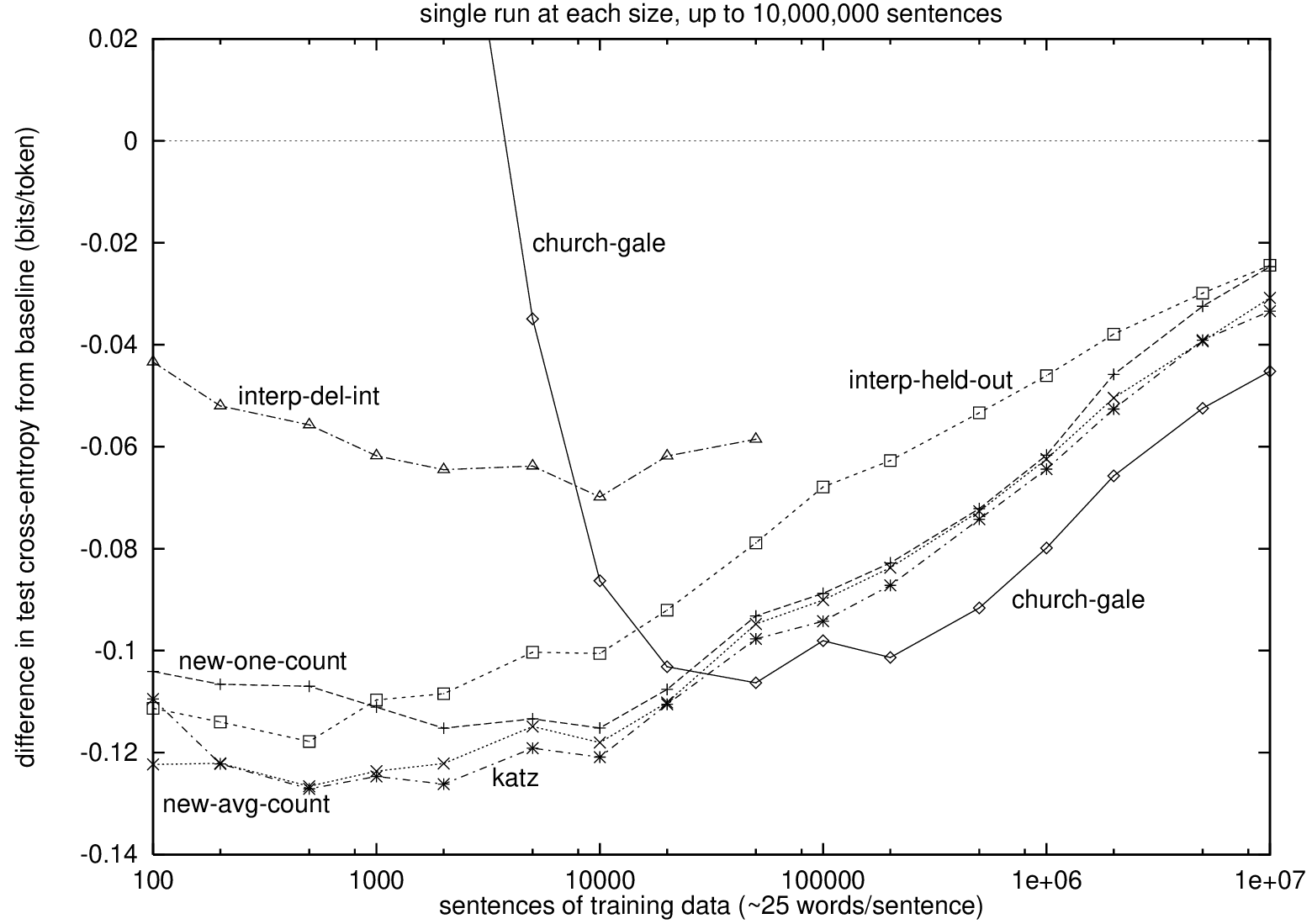,width=3in} $$
\caption{Bigram model on TIPSTER data;
	relative performance of various methods with respect
	to baseline; graphs on left display
	averages over ten runs for training sets up to 50,000 sentences, graphs
	on right display single runs for training sets up to 10,000,000 sentences;
	top graphs show all algorithms, bottom graphs zoom in on those methods
	that perform better than the baseline method}
	\label{fig:tip2}
\end{figure*}

\begin{figure*}[p]
$$ \psfig{figure=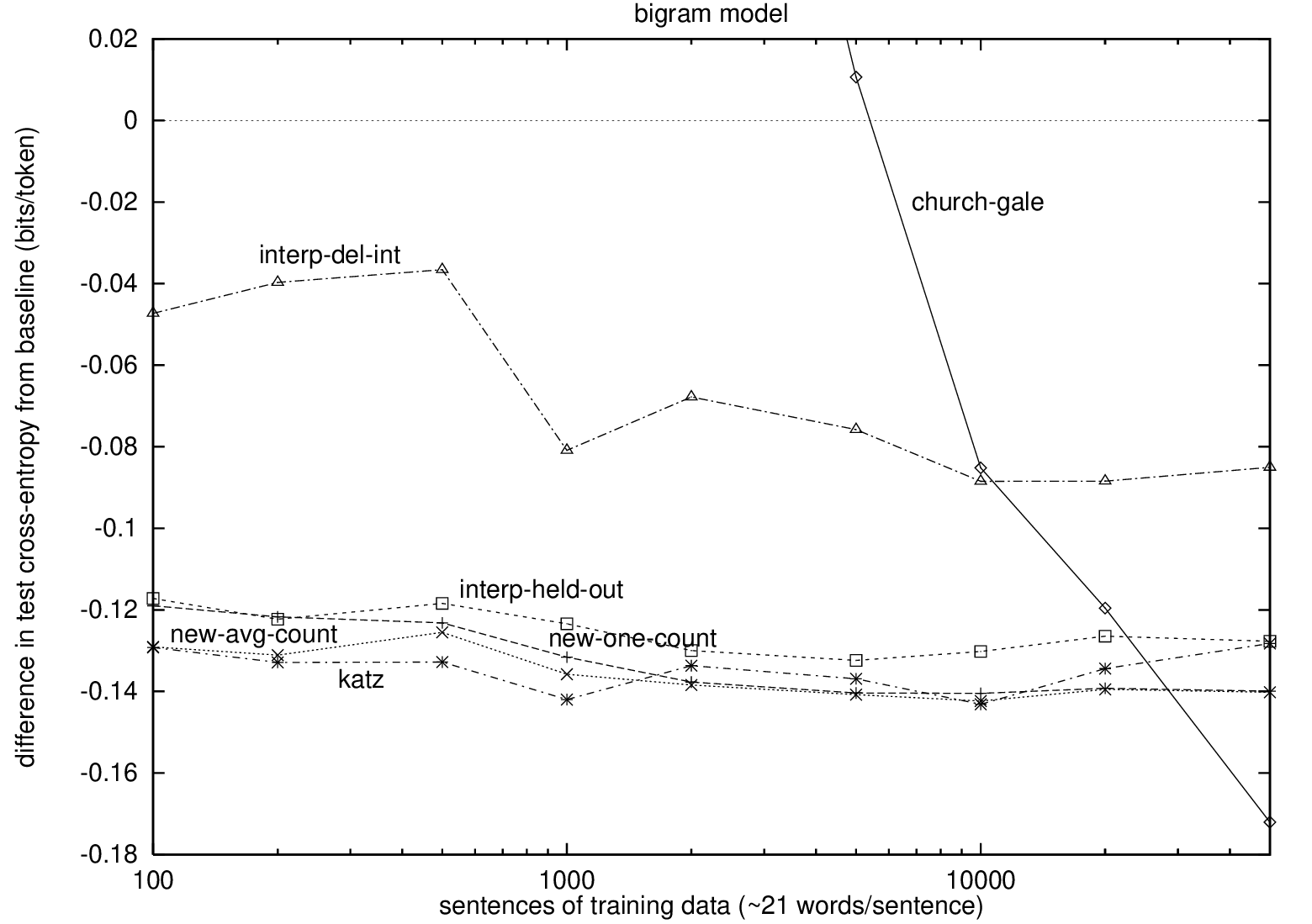,width=3in} \hspace{0.3in}
\psfig{figure=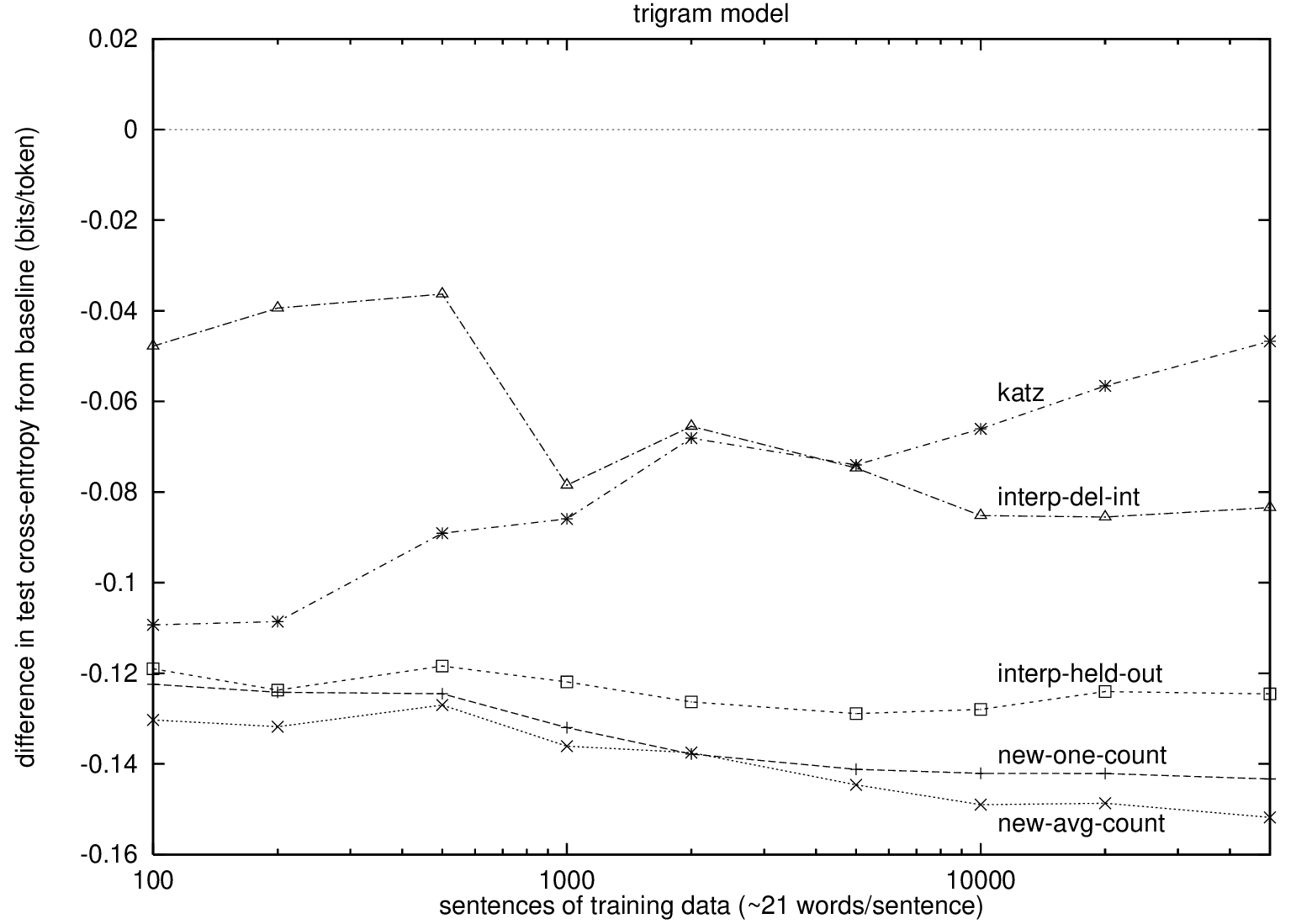,width=3in} $$
\caption{Bigram and trigram models on Brown corpus;
	relative performance of various methods with respect
	to baseline} \label{fig:brown}
\end{figure*}

\begin{figure*}[t]
$$ \psfig{figure=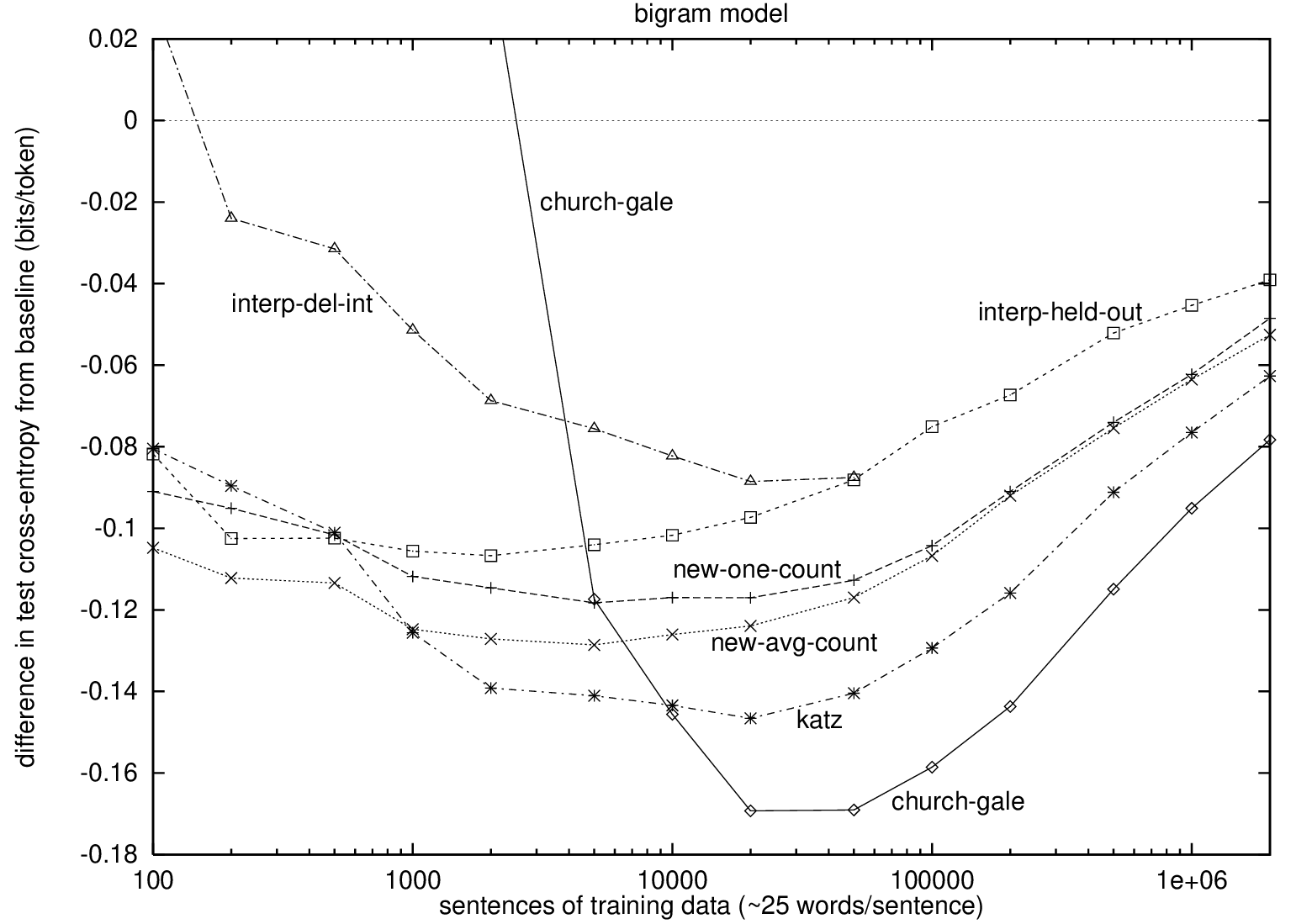,width=3in} \hspace{0.3in}
\psfig{figure=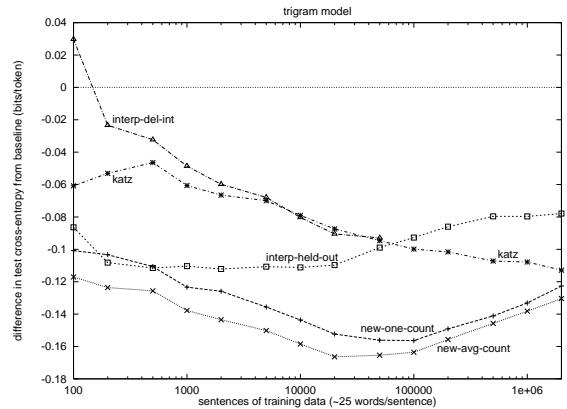,width=3in} $$
\caption{Bigram and trigram models on Wall Street Journal corpus;
	relative performance of various methods with respect
	to baseline} \label{fig:wsj}
\end{figure*}

\begin{figure*}[t]
$$ \psfig{figure=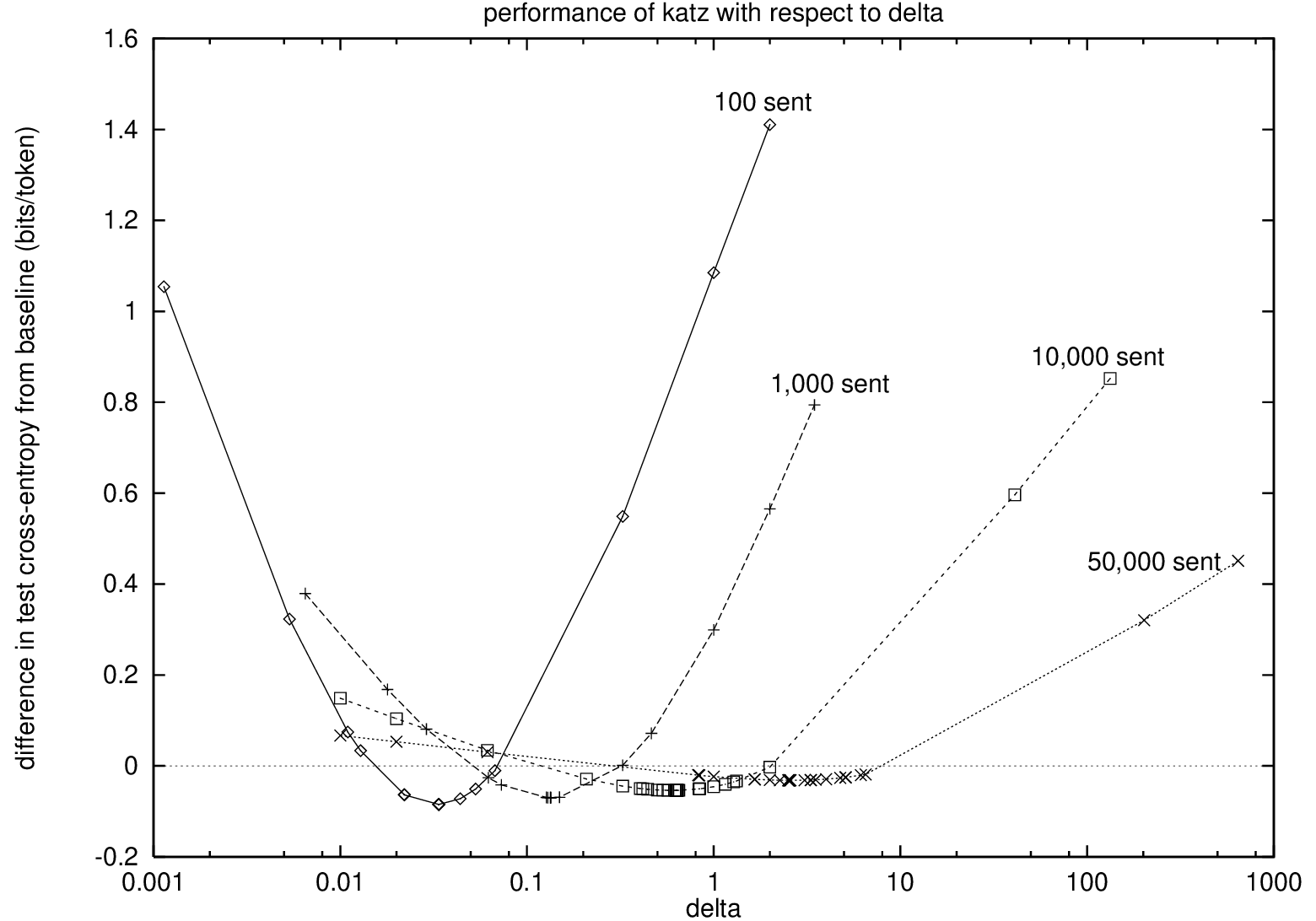,width=3in} \hspace{0.3in}
\psfig{figure=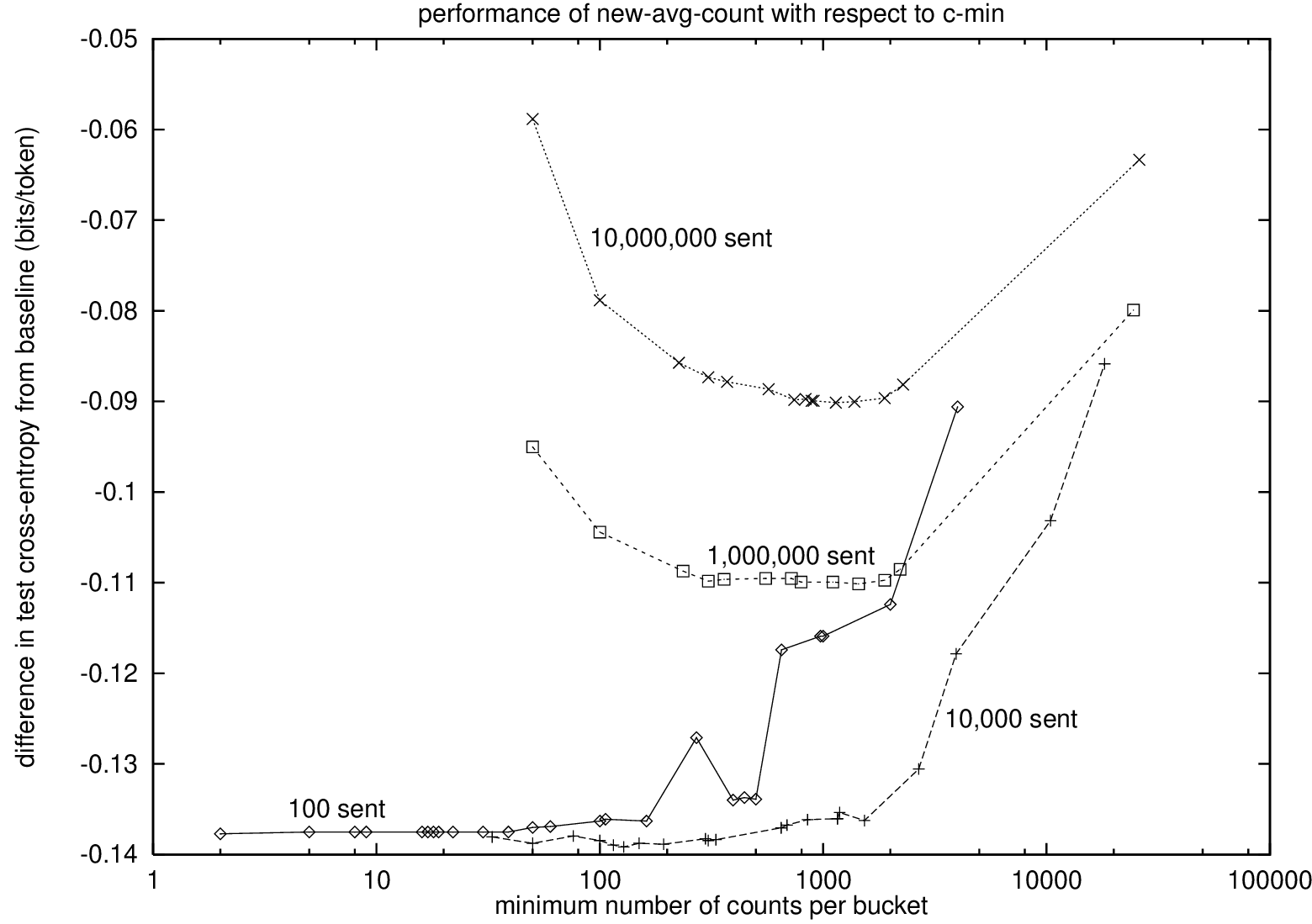,width=3in} $$
\caption{Performance of {\tt katz} and {\tt new-avg-count} with respect to parameters
	$\delta$ and $c_{\mbox{\protect\scriptsize min}}$, respectively}
	\label{fig:param}
\end{figure*}

\section{Results}

In Figure \ref{fig:base}, we display
the performance of the {\tt interp-baseline}
method for bigram and trigram models on TIPSTER, Brown, and the WSJ subset
of TIPSTER.  In Figures \ref{fig:tip3}--\ref{fig:wsj},
we display the
relative performance of various smoothing techniques with respect to
the baseline method on these corpora, as measured by difference
in entropy.  In the
graphs on the left of Figures \ref{fig:base}--\ref{fig:tip2}, each
point represents an average over ten runs; the error bars
represent the empirical standard deviation over these runs.
Due to resource limitations, we only performed multiple runs for
data sets of 50,000 sentences or less.  Each point on the graphs
on the right represents a single run, but we consider sizes up to the amount
of data available.  The graphs on the bottom of Figures
\ref{fig:tip3}--\ref{fig:tip2} are close-ups of the graphs above, focusing
on those algorithms that perform better than the baseline.
To give an idea of how these
cross-entropy differences translate to perplexity, each 0.014 bits
correspond roughly to a 1\% change in perplexity.

In each run except as noted below, optimal values for the parameters
of the given technique were searched for using Powell's search
algorithm as realized in {\it Numerical Recipes in C}
\cite[pp. 309--317]{Press:88a}.  Parameters were chosen to optimize the
cross-entropy of one of the development test sets associated with the
given training set.  To constrain the search, we searched only those
parameters that were found to affect performance significantly, as
verified through preliminary experiments over several data sizes.  For
{\tt katz} and {\tt church-gale}, we did not perform the parameter
search for training sets over 50,000 sentences due to resource
constraints, and instead manually extrapolated parameter values from
optimal values found on smaller data sizes.  We ran {\tt interp-del-int}
only on sizes up to 50,000 sentences due to time constraints.

From these graphs, we see that additive smoothing performs poorly and that
methods {\tt katz} and {\tt interp-held-out} consistently perform well.
Our implementation {\tt church-gale} performs poorly except on large
bigram training sets, where it performs the best.
The novel methods {\tt new-avg-count} and {\tt new-one-count} perform
well uniformly
across training data sizes, and are superior for trigram models.
Notice that while performance is relatively consistent across corpora,
it varies widely with respect to training set size and $n$-gram order.

The method {\tt interp-del-int} performs significantly
worse than {\tt interp-held-out}, though they differ only in the
data used to train the $\l$'s.  However, we delete one word at a time
in {\tt interp-del-int}; we hypothesize that deleting
larger chunks would lead to more similar performance.

In Figure \ref{fig:param}, we show how the values of the parameters $\d$
and $c\ss{min}$ affect the performance of methods {\tt katz} and
{\tt new-avg-count}, respectively, over several training data sizes.  Notice
that poor parameter setting can
lead to very significant losses in performance, and that optimal
parameter settings depend on training set size.

\begin{table}

$$ \begin{tabular}{|l|r|} \hline
Method & Lines \\ \hline
{\tt interp-baseline}\footnotemark & 400 \\ \hline
{\tt plus-one} & 40 \\ \hline
{\tt plus-delta} & 40 \\ \hline
{\tt katz} & 300 \\ \hline
{\tt church-gale} & 1000 \\ \hline
{\tt interp-held-out} & 400 \\ \hline
{\tt interp-del-int} & 400 \\ \hline
{\tt new-avg-count} & 400 \\ \hline
{\tt new-one-count} & 50 \\ \hline
\end{tabular} $$
\caption{Implementation difficulty of various methods in terms
of lines of C++ code} \label{tab:diff}
\end{table}

To give an informal estimate of the difficulty of implementation
of each method, in Table \ref{tab:diff}
we display the number of lines of C++ code in each
implementation excluding the core code common across techniques.
\footnotetext{
To implement the baseline method,
we just used the {\tt interp-held-out} code as
it is a special case.  Written anew, it probably would have been
about 50 lines.}

\section{Discussion}

To our knowledge, this is the first empirical comparison of
smoothing techniques in language modeling of such scope: no other
study has used multiple training data sizes, corpora, or has
performed parameter optimization.  We show that in order to completely
characterize the relative performance of two techniques, it
is necessary to consider multiple training set sizes and to try both
bigram and trigram models.  Multiple runs should
be performed whenever possible to
discover whether any calculated differences are
statistically significant.  Furthermore, we show that sub-optimal
parameter selection can also significantly affect relative performance.

We find that the two most widely used techniques, Katz smoothing
and Jelinek-Mercer smoothing, perform consistently well across
training set sizes for both bigram and trigram models, with Katz
smoothing performing better on trigram models produced from
large training sets and on bigram models in general.  These results
question the generality of the previous reference result concerning
Katz smoothing: \newcite{Katz:87a} reported that his method
slightly outperforms an unspecified version of Jelinek-Mercer smoothing
on a single training set of 750,000 words.  Furthermore, we show
that Church-Gale smoothing, which previously had not been compared
with common smoothing techniques, outperforms all existing methods
on bigram models produced from large training sets.  Finally,
we find that our novel methods {\it average-count\/} and {\it one-count\/}
are superior to existing methods for trigram models and perform well
on bigram models; method {\it one-count\/} yields marginally
worse performance but is extremely easy to implement.

In this study, we measure performance solely through the cross-entropy
of test data; it would be interesting to see how these cross-entropy
differences correlate with performance in end applications such
as speech recognition.  In addition,
it would be interesting to see whether these results extend to fields
other than language modeling where smoothing is used, such as
prepositional phrase attachment \cite{Collins:95a}, part-of-speech
tagging \cite{Church:88a}, and stochastic parsing \cite{Magerman:94a}.

\section*{Acknowledgements}

The authors would like to thank Stuart Shieber
and the anonymous reviewers for their comments on previous versions of
this paper.  We would also like to thank William Gale and Geoffrey
Sampson for supplying us with code for ``Good-Turing frequency estimation
without tears.''  This research
was supported by the National Science Foundation under
Grant No.\ IRI-93-50192 and Grant No.\ CDA-94-01024.
The second author was also supported by a National
Science Foundation Graduate Student Fellowship.

%

\end{document}